# Evaluating the Feasibility and Accuracy of Large Language Models for Medical History-Taking in Obstetrics and Gynecology

Dou Liu [1*], Ying Long [2*], Sophia Zuoqiu [3], Tian Tang [2*], Rong Yin [3*]

[1] Department of Industrial and Operation Engineering, University of Michigan, Ann Arbor, US

[2] Center for Reproductive Medicine, Department of Gynecology and Obstetrics, West China Second University Hospital, Sichuan University, Chengdu, China

[3] Department of Industrial Engineering, Sichuan University, Chengdu, China

[*]These authors contributed equally to this work.

## Abstract

Effective physician-patient communications in pre-diagnostic environments, and most specifically in complex and sensitive medical areas such as infertility, are critical but consume a lot of time and, therefore, cause clinic workflows to become inefficient. Recent advancements in Large Language Models (LLMs) offer a potential solution for automating conversational medical history-taking and improving diagnostic accuracy. This study evaluates the feasibility and performance of LLMs in those tasks for infertility cases. An AI-driven conversational system was developed to simulate physician-patient interactions with ChatGPT-4o and ChatGPT-4o-mini. A total of 70 real-world infertility cases were processed, generating 420 diagnostic histories. Model performance was assessed using F1 score, Differential Diagnosis (DDs) Accuracy, and Accuracy of Infertility Type Judgment (ITJ). ChatGPT-4o-mini outperformed ChatGPT-4o in information extraction accuracy (F1 score: 0.9258 vs. 0.9029, $p = 0.045$, $d = 0.244$) and demonstrated higher completeness in medical history-taking (97.58% vs. 77.11%), suggesting that ChatGPT-4o-mini is more effective in extracting detailed patient information, which is critical for improving diagnostic accuracy. In contrast, ChatGPT-4o performed slightly better in differential diagnosis accuracy (2.0524 vs. 2.0048, $p > 0.05$). ITJ accuracy was higher in ChatGPT-4o-mini (0.6476 vs. 0.5905) but with lower consistency (Cronbach's $\alpha = 0.562$), suggesting variability in classification reliability. Both models demonstrated strong feasibility in automating infertility history-taking, with ChatGPT-4o-mini excelling in completeness and extraction accuracy. In future studies, expert validation for accuracy and dependability in a clinical setting, AI model fine-tuning, and larger datasets with a mix of cases of infertility have to be prioritized.

**Keywords**

Large Language Models, Medical history-taking, Doctor-Patient Communication, AI-driven Healthcare, Obstetrics and Gynecology.

## 1. Problem Description

With the sociocultural pressures, lifestyle changes, and accelerated career pace of modern society, Infertility is becoming an prominent public health problem [1]. In the past decades, no significant changes in infertility independent from population growth and worldwide pregnancy declines [2]. Therefore, infertility remains a critical problem accompanied by a complicated process. Infertility not only brings emotional and financial burdens to those families but also challenges the hospitals' processing capability. Traditional procedures usually include pre-registration, initial screening, hormone testing, ultrasound, medical history analysis, and subsequent design of individualized treatment plans, etc. Medical history-taking is one of the most time-consuming procedures; it takes a relatively long period for physicians to conduct introductory interviews and record the patient's information. Additionally, it has been suggested that 60-80% of diagnoses are made through medical history-taking alone[3], [4]. Thus, optimizing and decreasing unnecessary repetitive and easy work and improving efficiency tend to be key challenges that need to be addressed.

## 2. Related Research

However, as Artificial Intelligence (AI) takes a giant part of current industries, it seems to offer a new solution to this key problem. Large Language Models (LLMs) have been proven to have fabulous professional medical knowledge in multiple languages [5], [6] and have proven capable of highly accurate single-turn medical question-answering [7]. The use of AI Physician Agent (AI Physician Agent) for initial consultation and information collection is gaining attention, with LLMs' conversational capabilities tailored to domains outside clinical medicine [8]. This shift could save human physicians a great deal of time in the initial diagnosis since they can directly check the report generated by the agent and reference the recommendations (Necessary inspection items or assessment of associated risks) for those novice physicians. In real-world practice, some hospitals have already adopted chatbots to pre-collect patients' information. In the previous study, a team tried to construct a model specializing in diagnostic dialogue, and it showed a significantly higher differential diagnosis accuracy than primary care physicians [9]. Therefore, building on the background discussed above, this study seeks to evaluate the capability of AI Physician Agents in conducting history-taking consultations, generating patient records, and handing out differential diagnoses, specifically for infertility cases. To validate the system's effectiveness and feasibility, we have developed a real-world conversational simulation environment featuring two AI agents and utilizing hospital case data. By integrating AI physicians into outpatient settings, consultation times may be shortened, healthcare efficiency may be improved, and the burden on healthcare workers may be reduced: all while maintaining diagnostic accuracy.

## 3. Methodology

### 3.1 Data Source

The dataset was collected from West China Second University Hospital and was previously approved by the Institutional Review Board (IRB2023-278). For this pilot study, 70 cases were randomly selected from the database.

### 3.2 Study Design

A fully automated conversational system consisting of two AI agents was developed, each designed to simulate the roles of a physician and a patient. Upon activation, the physician agent initiates the interaction by posing the first question to the patient agent. This marks the beginning of a virtual medical consultation, lasting through to when a concluding diagnosis and a complete medical record have been developed by the physician agent. For the physician, the AI agent is programmed to act as an expert in infertility, having consultations with high clinical accuracy. Our prompt highlighted that the agent should focus on diagnosis, not simply collecting the information. To ensure the information gathered during the interaction remains focused and clinically relevant, the physician agent's prompt includes a structured template outlined in Table 1.

On the patient's side, the AI agent is programmed to emulate the behavior of an average individual with no formal medical training. Based on the specific case details embedded within its prompt, the patient agent responds to the

Liu, Long, Zuoqiu, Tang, Yinphysician's inquiries by providing accurate and contextually appropriate answers or indicating a lack of information when the queried details are not present in the case vignette. This design ensures the interaction remains realistic, simulating the dynamics of real-world medical consultations while enabling a robust evaluation of the system's ability to extract and synthesize clinically relevant information.

Table 1: Structured template of the Categories and Check Point Parameters

| Category | Check Point Parameters | | |
|---|---|---|---|
| **Basic Information** | Age | History of Present Condition | Medications |
| | Chief Complaint | Past Medical History | Surgical History |
| **Infertility History** | Duration of Infertility | Past Pregnancies Number | Abortions Number |
| | Menstrual Cycle | Kids Alive Number | Abortion Type |
| | Menstrual Duration | Delivery Type | |
| **Past Examination** | Histogenesis Examination of Abortion | Anti-Müllerian hormone | Semen Analysis for Male Partner |
| | Hysterosalpingography | Intrauterine insemination | Tubal Flushing |
| | Hysteroscopy & Laparoscopy | In Vitro Fertilization | |

After constructing the automated diagnostic system, we conducted three diagnostic evaluations for each case using ChatGPT-4o and ChatGPT-4o-mini, resulting in a total of six diagnostic histories per case. This study processed 70 distinct infertility cases, theoretically generating 420 diagnostic histories in total. We employed a range of quantitative evaluation metrics to assess each model's feasibility and performance. Both within-model consistency (assessing the reliability of repeated diagnoses within the same model) and between-model analysis (comparing the diagnostic outputs across models) were systematically analyzed.

### 3.3 Completeness of medical history-taking

Completeness reflects the Agent's capacity to complete the consultant form. To evaluate the completeness of the physician AI agent's questions, a percentage grading system was adopted. With this, a transparent and objective expression of AI's thoroughness in including the key elements of medical history could be achieved, and a uniform criterion for evaluation could be determined. The completeness score is computed in equation (1) below:

$$Completeness\ Score = \frac{Number\ of\ Covered\ Points}{Total\ Key\ Points} \times 100\% \tag{1}$$

### 3.4 Accuracy of Record Extraction

To quantitatively evaluate the system's ability to elicit and accurately document information during medical history-taking, we propose using the F1 score as the evaluation metric. First, each discrete piece of information in the original vignette is defined as a single "point" (for example, the statement "She has had a headache for two weeks" is counted as one point). Then, we compare the generated record with the original vignette to identify which points have been correctly captured. *Precision* is calculated as the ratio of correctly documented points in the generated record to the total number of points present in the generated record. *Recall* is defined as the ratio of the number of correctly documented points to the total number of points in the original vignette. Finally, the F1 score is computed as the harmonic mean of precision and recall:

$$F1\ Score = 2 \times \frac{Precision \times Recall}{Precision + Recall} \tag{2}$$

## 3.5 Accuracy of Differential Diagnosis and Infertility Type Judgement

As the core outputs, the Differential Diagnosis (DDs) and the Infertility Type Judgment (ITJ) are evaluated using distinct scoring systems. For DDs, we employed a 3-point Likert scale where: 1) A score of 1 indicates that the offered differential diagnoses bear no relevance to the gold standard. 2) A score of 2 indicates that the differential diagnoses exhibit partial relevance, with some elements aligning with the gold standard. 3) A score of 3 indicates that the generated differential diagnoses comprehensively cover all aspects presented in the original case. We adopted the 3-point Likert scale to provide a simple, yet clinically meaningful way for experts to evaluate the alignment of AI-generated differential diagnoses with gold standard cases. This approach captures nuanced partial correctness that binary scoring may miss. For ITJ, a binary scoring system was applied, where a score of 1 denoted a correct judgment (True) and a score of 0 denoted an incorrect judgment (False). In this study, we use the t-test, Cronbach's Alpha, and Cohen's d to analyze all the data collected. Python (version 3.12.1) was used to conduct all the tests. A statistical significance level of α=0.05 was applied.

## 4. Results

All cases in this study focused specifically on infertility, with the majority of diagnoses attributed to female factors, such as tubal disorders and ovarian reserve deficiencies. The performance of two representative LLMs is compared, i.e., ChatGPT-4o-mini and ChatGPT-4o. In terms of medical history-taking completeness, ChatGPT-4o-mini performed better than ChatGPT-4o, suggesting increased thoroughness in elicitation and documentation of patient histories, potentially contributing to fuller clinical assessments. Completeness, however, is not a reflection of extraction accuracy. The quantitative evaluation of model performance, including F1 score, Differential Diagnosis (DDs) Accuracy, and Infertility Type Judgment (ITJ) Accuracy. Among the 420 diagnostic histories generated, ChatGPT-4o-mini demonstrated a significantly higher F1 score (0.9258) compared to ChatGPT-4o (F1 score = 0.9029, p=0.045). Despite this difference, both models performed well in extracting medical histories, with F1 scores exceeding 0.9. Furthermore, ChatGPT-4o-mini showed greater stability with lower standard deviation (0.0636 vs. 0.0981). For diagnosis accuracy, ChatGPT-4o slightly outperformed ChatGPT-4o-mini (2.0524 vs. 2.0048), but the difference was not statistically significant, suggesting comparable performance in generating differential diagnoses. In terms of ITJ accuracy, ChatGPT-4o-mini achieved a higher accuracy (0.6476) compared to ChatGPT-4o (0.5905), with a smaller standard deviation (0.3447 vs. 0.3979), indicating improved reliability in classification accuracy. However, the difference in DDs and ITJ accuracy was insignificant. These findings suggest that while ChatGPT-4o-mini exhibits slightly superior performance in information extraction and classification accuracy; both models demonstrate comparable effectiveness in generating differential diagnoses. Overall, the results indicate that both models perform well in processing medical histories, with ChatGPT-4o-mini providing more stable output.

In terms of consistency, we analyzed Cronbach's alpha across three key evaluation dimensions to estimate the reliability of AI output produced through them. In accordance with our analysis, the majority of the Cronbach's alpha values were above the widely accepted threshold for reliability and thus, high on internal consistency in model performance. Regarding accuracy for ITJ in ChatGPT-4o-mini, with a comparatively low value for Cronbach's alpha (α = 0.562). That indicates a lack of stability and replicability in ITJ classification across cases and perhaps reflects variable decision-making or inconsistency in case-data interpretation in the model. That lower level of consistency may occur for a range of reasons, including limitations in training data for the model, challenge in handling uncertain case descriptions, or variability in AI-based classification of infertility types per se. As consistent classification is crucial in medical diagnostics, these findings identify areas in which refinement and fine-tuning are necessary to render AI-support for infertility diagnostics stronger and more reliable.

## 5. Discussion

This study proposed to evaluate the performance and viability of ChatGPT-4o and ChatGPT-4o-mini in taking medical histories and diagnosing infertility cases. As per the findings, ChatGPT-4o-mini performed with a better F1 score, with a reflection of heightened accuracy in extracting medical information pertinent to cases. Besides, ChatGPT-4o-mini exhibited a high completeness in taking medical histories. However, even with its performance, ChatGPT-4o exhibited a marginally high accuracy in differential diagnosis, but not one that reached a level of statistical significance, and therefore, an equivalent effectiveness in thinking in a clinical manner. On the other hand, even with a high accuracy in infertility type judgment (ITJ), the inner reliability in its classification is not high. Therefore, the consistency of classification needs to be further addressed in future improvements.

Previous studies have analyzed AI-powered diagnostics in pre-consultation settings with successful performance in a variety of medical specialties. For instance, specific groups have designed and evaluated custom AI algorithms and have seen that their systems performed better in outpatient pediatrics when having access to the same medical background [10]. In most recent studies, in addition, the developed system performed a high level of differential diagnostic accuracy in terms of specialties and practitioners [9]. In specific diseases, for instance, Parkinson's, AI assistants' satisfactory level of performance in diagnostics was stressed [11], in a similar level of clinical thinking and cannot-miss diagnoses included [12]. This study demonstrated the potential of using AI-powered chatbots in enhancing medical history-taking efficiency in infertility cases. By offering information extraction and pre-diagnostic evaluation, ChatGPT-4o-mini may simplify much work for medical professionals, allowing them to spend more time in high-order medical thinking and caring for patients. Besides, high completeness in medical documentation through ChatGPT-4o-mini shows that AI can make medical documentation less variable and less prone to omission and errors in patient files. Instability in ITJ classification, however, shows a critical disadvantage, and AI-powered classification must therefore be cautiously analyzed and validated through clinicians' feedback before incorporation in patient care management planning. With a narrow margin in accuracy in differential diagnoses between both AI-powered chatbots, both models can function for AI-powered differential diagnoses in cases of infertility.

One primary challenge in deploying AI-driven medical consultations is the development of suitable prompts. While domain experts acknowledged the capabilities of the AI system, they emphasized the need for improved clinical reasoning—particularly in cases with complex diagnostic factors or nuanced differential diagnoses. For example, the relatively low ITJ accuracy and consistency may stem from ambiguity and overlap among certain infertility subtypes, especially in borderline or multifactorial cases. Future efforts should prioritize clearer classification criteria and incorporate expert-reviewed gold standard tailored to enhance accuracy and reliability. This study did not include direct evaluations from medical professionals. Clinician feedback will be essential for assessing both the realism of the AI-generated dialogues and the clinical validity of their outputs. With additional fine-tuning using a targeted training dataset, the AI's diagnostic reasoning and specificity could be substantially improved. To address these challenges, future research will focus on developing a fine-tuned model trained on a larger, more diverse dataset to optimize diagnostic performance. Further enhancements may include integrating reinforcement learning with clinician-in-the-loop feedback to continuously refine decision-making, thereby increasing the system's reliability, consistency, and clinical utility. Real-world deployment of LLMs in clinical settings may raise potential ethical and legal concerns, particularly related to HIPAA compliance and patient data privacy. Commercial LLM providers may store data during inference, posing potential risks unless systems are deployed within secure environments. Future work should consider privacy deployment strategies, such as fine-tuned open-source models hosted within hospital infrastructure. Finally, alternative NLP models like BioGPT and ClinicalBERT offer domain-specific advantages, future studies may explore hybrid architectures that combine these strengths—such as integrating domain-specific models into retrieval-augmented generation (RAG) pipelines—to further improve clinical relevance and performance.